\documentclass[10pt,twocolumn,a4paper]{article}

\usepackage[utf8]{inputenc}
\usepackage[T1]{fontenc}
\usepackage[margin=1in]{geometry}
\usepackage{amsmath,amssymb,amsfonts,amsthm}

\usepackage{graphicx}
\graphicspath{{figures/}}
\usepackage{caption}
\usepackage{subcaption}

\usepackage{booktabs}
\usepackage{algorithm}
\usepackage{algpseudocode}

\usepackage[numbers]{natbib}
\usepackage{hyperref}
\hypersetup{
    colorlinks=true,
    linkcolor=black,
    citecolor=blue,
    urlcolor=blue
}

\title{\textbf{Manifold Decoders: A Framework for Generative Modeling from Nonlinear Embeddings}}

\author{
    \normalsize Riddhish Thakare\thanks{Corresponding Author: riddhishthakare@ufl.edu} \textsuperscript{1} \and
    \normalsize Kingdom Mutala Akugri \textsuperscript{1}
}

\date{
    \textsuperscript{1}Department of Computer \& Information Science \& Engineering, University of Florida \\
    \textsuperscript{1}Department of Computer \& Information Science \& Engineering, University of Florida \\
}

\begin{document}

\twocolumn[
\begin{@twocolumnfalse}
\maketitle

\begin{abstract}
\noindent
Classical nonlinear dimensionality reduction (NLDR) techniques like t-SNE, Isomap, and LLE excel at creating low-dimensional embeddings for data visualization but fundamentally lack the ability to map these embeddings back to the original high-dimensional space. This one-way transformation limits their use in generative applications. This paper addresses this critical gap by introducing a systematic framework for constructing neural decoder architectures for prominent NLDR methods, enabling bidirectional mapping for the first time. We extend this framework by implementing a diffusion-based generative process that operates directly within these learned manifold spaces. Through experiments on the CelebA dataset, we evaluate the reconstruction and generative performance of our approach against autoencoder and standard diffusion model baselines. Our findings reveal a fundamental trade-off: while the decoders successfully reconstruct data, their quality is surpassed by end-to-end optimized autoencoders. Moreover, manifold-constrained diffusion yields poor-quality samples, suggesting that the discrete and sparse nature of classical NLDR embeddings is ill-suited for the continuous interpolation required by generative models. This work highlights the inherent challenges in retrofitting generative capabilities onto NLDR methods designed primarily for visualization and analysis.
\end{abstract}

\vspace{1em}
\begin{center}
\textbf{Keywords:} Manifold Learning, Generative Models, Diffusion Models, Nonlinear Dimensionality Reduction
\end{center}

\vspace{2em}
\end{@twocolumnfalse}
]


\section{Introduction}
High-dimensional data analysis and visualization constitute fundamental challenges in machine learning, where nonlinear dimensionality reduction (NLDR) techniques have proven instrumental in discovering low-dimensional embeddings that preserve essential structural properties of complex datasets. These methods, encompassing techniques such as t-distributed Stochastic Neighbor Embedding (t-SNE) ~\cite{Maaten2008TSNE} , Isometric Mapping (Isomap) ~\cite{Tenenbaum2000Isomap} , Locally Linear Embedding (LLE) ~\cite{roweis2000lle}  and Laplacian Eigenmaps ~\cite{Belkin2003Laplacian} excel at revealing intrinsic data manifolds and facilitating interpretable visualizations of high-dimensional phenomena. However, a critical architectural limitation pervades the entire class of traditional NLDR methods: they inherently lack reconstruction capabilities, operating as one-way transformations that map from high-dimensional input spaces to low-dimensional embeddings without providing mechanisms for inverse mapping.

This fundamental asymmetry severely constrains the applicability of NLDR techniques in generative modelling, data synthesis, and interactive exploration scenarios where bidirectional transformations are essential. Unlike autoencoders, which explicitly incorporate decoder architectures during training, classical manifold learning approaches such as t-SNE, Uniform Manifold Approximation and Projection (UMAP) ~\cite{McInnes2018UMAP} , and diffusion maps optimize embeddings through eigen decomposition, neighbourhood preservation, or probabilistic formulations that do not naturally yield invertible mappings. Consequently, despite their superior performance in preserving local neighbourhood structures and global topological properties, these methods remain confined to analysis and visualization tasks.

This work addresses the reconstruction gap in NLDR methods by developing specialized decoder architectures that enable bidirectional mapping between high-dimensional data and learned manifold representations. Our approach systematically constructs decoders for prominent NLDR techniques, including t-SNE, Isomap, and LLE, enabling these methods to participate in generative workflows for the first time. We extend this framework to manifold diffusion processes, demonstrating how NLDR-equipped decoders can facilitate novel sample generation through diffusion operations performed directly in the learned embedding space rather than the original high-dimensional domain. Through evaluation against autoencoder baselines on the Celeb-A dataset, we characterize the fundamental trade-offs between manifold-based and traditional reconstruction approaches, providing empirical insights into their complementary strengths for different data types and generative tasks. 
\section{Related Work}
\subsection{Manifold Learning and Dimensionality Reduction}

The field of nonlinear dimensionality reduction has evolved significantly since the pioneering work of the early 2000s, with several landmark methods establishing the theoretical and practical foundations for manifold discovery in high-dimensional data. Locally Linear Embedding (LLE)  introduced the principle of preserving local linear relationships by reconstructing each data point as a weighted combination of its nearest neighbors, subsequently finding a low-dimensional embedding that maintains these reconstruction weights. ISOMAP extended classical Multidimensional Scaling by replacing Euclidean distances with geodesic distances computed along the data manifold, enabling the preservation of intrinsic geometric structure in non-convex manifolds.

Spectral methods have provided another powerful paradigm, with Laplacian Eigenmaps leveraging graph Laplacian eigendecomposition to discover embeddings that respect local neighborhood structures while maintaining global connectivity patterns. The probabilistic approach exemplified by t-distributed Stochastic Neighbor Embedding (t-SNE)  revolutionized high-dimensional data visualization by optimizing probabilistic similarities between point pairs, employing heavy-tailed distributions in the embedding space to alleviate crowding problems inherent in lower dimensions. More recently, Uniform Manifold Approximation and Projection (UMAP) has emerged as a computationally efficient alternative that combines topological data analysis with Riemannian geometry to achieve both local structure preservation and global connectivity. 

\subsection{The Reconstruction Problem in Manifold Learning }

A fundamental limitation shared across traditional NLDR methods is their inherent inability to perform inverse transformations from embedding space back to the original high-dimensional domain. This asymmetry stems from the mathematical formulations underlying these techniques: spectral methods rely on eigendecomposition of similarity or distance matrices, optimization-based approaches like t-SNE employ non-convex objective functions without explicit inverse mappings, and neighborhood-preserving methods focus solely on forward embedding computation. Early attempts to address this limitation include the work of Bengio et al. ~\cite{Bengio2003ManifoldParzen} , who proposed learning approximate inverse mappings using multilayer perceptrons trained on embedded-original data pairs, though this approach requires separate training phases and cannot guarantee consistency with the original manifold structure. 

\subsection{Autoencoders and Representation Learning}

In contrast to classical NLDR methods, autoencoder architectures ~\cite{Hinton2006Autoencoder} explicitly incorporate both encoding and decoding components, enabling bidirectional transformations between input and latent spaces. Variational Autoencoders ~\cite{pu2016variational} further extended this paradigm by introducing probabilistic latent representations, while more recent developments in deep generative modeling have demonstrated the power of learned representations for synthesis tasks. However, these approaches typically optimize reconstruction objectives that may not preserve the specific geometric and topological properties prioritized by manifold learning methods, potentially sacrificing manifold structure for reconstruction fidelity. 

\subsection{Diffusion Models and Manifold-Based Generation}

The recent success of diffusion probabilistic models ~\cite{Ho2020DDPM} has demonstrated the effectiveness of iterative denoising processes for high-quality sample generation. While most diffusion approaches operate directly in data space or learned autoencoder latents, several works have explored diffusion in alternative geometric spaces ~\cite{debortoli2022riemannian}. Our work contributes to this emerging direction by investigating diffusion processes constrained to manifolds discovered by classical NLDR methods. 
\section{Materials and Methods}
\subsection{Dataset and Preprocessing }
We conduct our experiments on the CelebA dataset ~\cite{liu2015deep}, a large-scale facial attribute dataset comprising over 200,000 high-resolution celebrity images with 40 binary attribute annotations. To ensure computational tractability while maintaining sufficient complexity for manifold learning analysis, we employ a stratified subset of 10,000 images that preserves the original attribute distribution. All images undergo standardized preprocessing: resizing to 64×64 pixels using bicubic interpolation, conversion to RGB format, and normalization to the range [-1, 1] through min-max scaling. This resolution provides sufficient detail for facial structure analysis while enabling efficient computation across multiple manifold learning techniques. 

\subsection{Manifold Learning Pipeline  }
We systematically evaluate four prominent nonlinear dimensionality reduction (NLDR) methods, each representing a distinct theoretical approach to manifold discovery. 

\textbf{Locally Linear Embedding (LLE)}~\cite{roweis2000lle} preserves local neighborhood relationships through linear reconstruction weights, configured with $k = 12$ nearest neighbors and embedding dimension $d = 50$. 

\textbf{ISOMAP}~\cite{Tenenbaum2000Isomap} maintains geodesic distance relationships via shortest-path computations on the $k$-nearest neighbor graph ($k = 10$), projecting data into the same $50$-dimensional latent space. 

\textbf{Laplacian Eigenmaps (LE)}~\cite{Belkin2003Laplacian} employs spectral decomposition of the graph Laplacian constructed from Gaussian similarity kernels with bandwidth $\sigma = 1.0$ and $k = 15$ neighbors, yielding embeddings that preserve local graph connectivity. 

Finally, \textbf{t-SNE}~\cite{Maaten2008TSNE} optimizes probabilistic neighborhood relationships using perplexity $= 30$ and learning rate $= 200$, with early exaggeration applied for the first $250$ iterations to promote cluster separation.

\vspace{0.5em}
The choice of $50$-dimensional embeddings across all methods enables direct comparison while providing sufficient representational capacity for complex facial variations. Each method processes the flattened $12{,}288$-dimensional input vectors ($64 \times 64 \times 3$), applying its respective optimization procedure to discover the underlying manifold structure within the facial image distribution.

\subsection{Neural Decoder Architecture   }

To enable reconstruction from manifold embeddings, we design a unified decoder architecture adaptable to the output characteristics of each NLDR method. The decoder follows a progressive upsampling strategy, transforming the 50-dimensional manifold coordinates through a sequence of transposed convolutions. The architecture begins with a fully connected layer that projects the embedding to a 512-dimensional feature space, subsequently reshaped to 8×8×8 spatial dimensions.

Four successive transposed convolutional layers progressively increase spatial resolution while decreasing feature depth: the first layer (stride=2, kernel=4) expands to 16×16×128, followed by 32×32×64, then 64×64×32, and finally 64×64×3 for RGB output. Each intermediate layer incorporates batch normalization ~\cite{ioffe2015batch} to stabilize training dynamics and ReLU activation to introduce nonlinearity. The final layer employs tanh activation to constrain outputs to the normalized pixel range [-1, 1], ensuring consistency with the input preprocessing scheme. 
\begin{figure*}[h!]
    \centering
    \includegraphics[width=\textwidth]{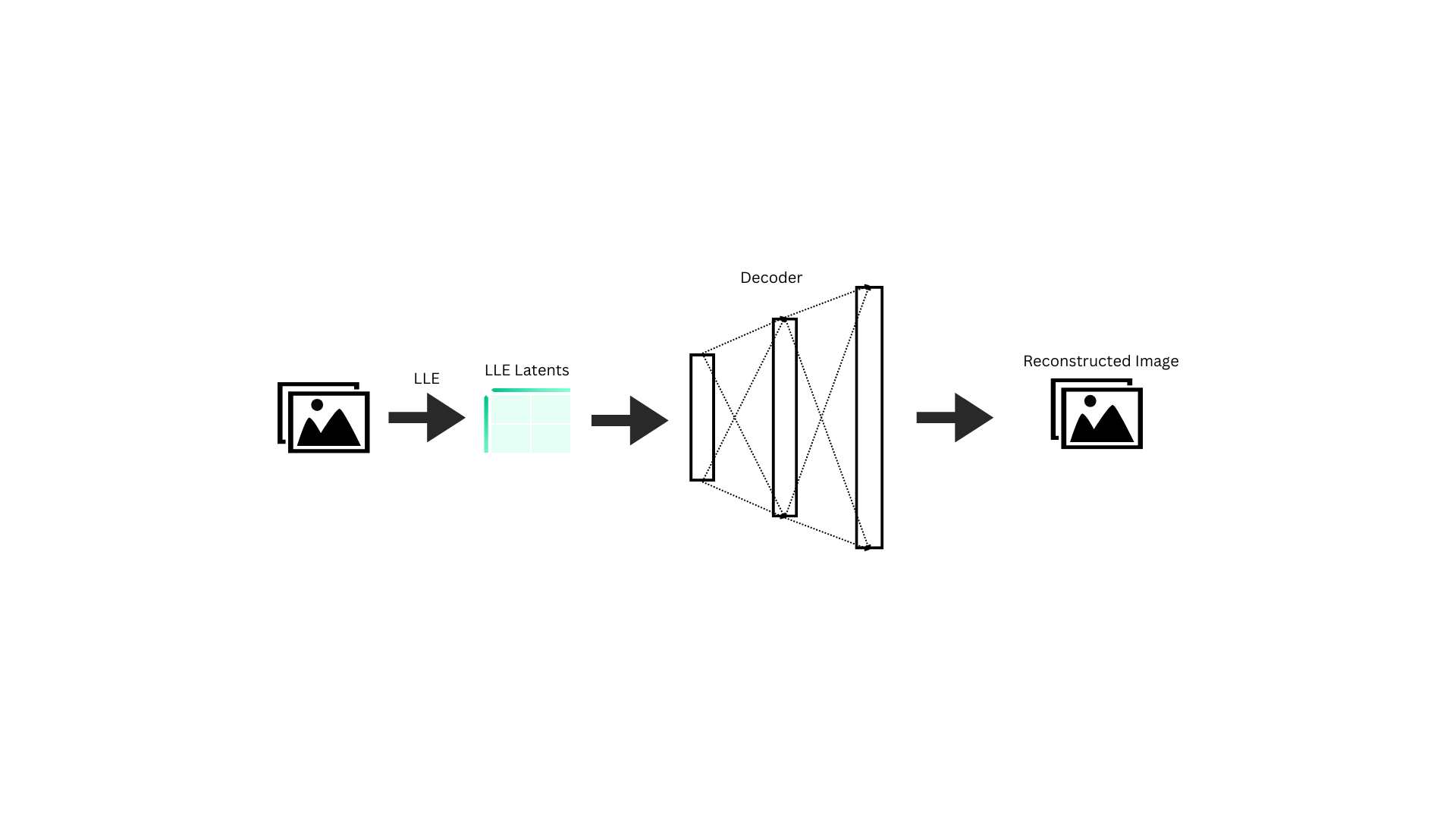}
    \caption{Overview of the manifold learning–based decoder framework.}
    \label{fig:placeholder}
\end{figure*}

\subsection{Decoder Training Framework}

The decoder training procedure employs a multi-component loss function designed to optimize both pixel-level reconstruction fidelity and perceptual quality. The total objective combines the Mean Squared Error (MSE) reconstruction loss with a perceptual component as follows:
\begin{equation}
\mathcal{L}_{\text{total}} = \lambda_{\text{MSE}} \, \mathcal{L}_{\text{MSE}} + \lambda_{\text{perceptual}} \, \mathcal{L}_{\text{perceptual}},
\end{equation}
where
\begin{equation}
\mathcal{L}_{\text{MSE}} = \left\lVert I_{\text{original}} - I_{\text{reconstructed}} \right\rVert_2^2
\end{equation}
provides pixel-wise supervision, and $\mathcal{L}_{\text{perceptual}}$ computes feature-space distances using pre-trained VGG-16 representations~\cite{tammina2019transfer} to preserve semantic content. Empirically, we set $\lambda_{\text{MSE}} = 1.0$ and $\lambda_{\text{perceptual}} = 0.1$ based on validation across manifold types.

Training employs the Adam optimizer with learning rate $\alpha = 2\times10^{-4}$, momentum parameters $\beta_1 = 0.9$, $\beta_2 = 0.999$, and weight decay $1\times10^{-4}$ to mitigate overfitting. Each decoder is trained for $50$ epochs with batch size $64$, incorporating a cosine annealing learning rate schedule to improve convergence stability.

Given the fixed nature of NLDR embeddings (computed once prior to decoder training), we apply coordinate dropout with probability $0.1$ to improve robustness to manifold perturbations and enhance generalization to interpolated regions of the embedding space.

Validation is conducted on a held-out $20\%$ subset using reconstruction metrics including MSE, Peak Signal-to-Noise Ratio (PSNR), and Structural Similarity Index (SSIM) to assess both quantitative accuracy and perceptual quality. Each decoder architecture is tuned for its corresponding NLDR method, accounting for the geometric properties preserved by LLE’s local linearity, ISOMAP’s geodesic distances, Laplacian Eigenmaps’ spectral structure, and t-SNE’s probability distributions.

\subsection{Manifold Diffusion Process }

To evaluate the generative capabilities enabled by our decoder framework, we implement a diffusion-based sampling procedure that operates directly within the learned manifold spaces rather than the original high-dimensional image domain. This approach leverages the geometric structure discovered by NLDR methods while enabling controlled sample generation through iterative denoising.

We first establish a baseline using a standard Denoising Diffusion Probabilistic Model (DDPM)~\cite{Ho2020DDPM} trained on grayscale CelebA images resized to $32\times32$ pixels for computational efficiency. The baseline employs a U-Net architecture with attention mechanisms, trained for $1000$ diffusion timesteps using a cosine noise schedule and DDPM’s reverse process for sample generation.

For manifold-based diffusion, we adapt the diffusion framework to operate within the $50$-dimensional embedding spaces produced by each NLDR method. The forward diffusion process gradually corrupts manifold coordinates $\mathbf{x}_0$ through Gaussian noise injection:
\begin{equation}
q(\mathbf{x}_t \mid \mathbf{x}_{t-1}) = \mathcal{N}\!\left(\mathbf{x}_t; \sqrt{1 - \beta_t}\,\mathbf{x}_{t-1},\, \beta_t \mathbf{I}\right),
\end{equation}
where $\beta_t$ follows a linear noise schedule from $\beta_1 = 1\times10^{-4}$ to $\beta_T = 0.02$ over $T = 1000$ timesteps.  
The reverse denoising process employs a lightweight MLP with three hidden layers of $256$ units each, predicting the noise component $\boldsymbol{\epsilon}_\theta(\mathbf{x}_t, t)$ given noisy manifold coordinates $\mathbf{x}_t$ and timestep $t$.

Critically, this manifold diffusion approach constrains sample generation to regions with meaningful decoder support, as the denoising network learns to generate coordinates consistent with the original NLDR embedding distribution. Generated manifold coordinates are subsequently decoded through the corresponding trained decoder to produce final image reconstructions, enabling assessment of both manifold interpolation quality and decoder robustness to novel latent representations.

The manifold diffusion training employs the standard DDPM objective:
\begin{equation}
\mathcal{L}_{\text{diffusion}} = 
\mathbb{E}_{t,\,\mathbf{x}_0,\,\boldsymbol{\epsilon}}
\!\left[\left\lVert 
\boldsymbol{\epsilon} - 
\boldsymbol{\epsilon}_\theta(\mathbf{x}_t, t)
\right\rVert_2^2\right],
\end{equation}
optimized using the Adam optimizer with learning rate $\alpha = 1\times10^{-4}$ for $100$ epochs.  
This framework enables systematic evaluation of how different manifold geometries—local versus global, spectral versus probabilistic—influence the quality and diversity of generated samples when combined with learned reconstruction mappings.

\section{Results}
\subsection{Manifold Learning Decoder Evaluation}
We systematically evaluate decoder performance across four prominent NLDR techniques using standardized experimental protocols. All experiments employ 25,000 CelebA images resized to 64×64 RGB format, enabling direct comparison of reconstruction quality across different manifold learning paradigms. 
\begin{table}[h!]
\centering
\caption{Comparison of reconstruction losses across methods.}
\label{tab:loss_comparison}
\begin{tabular}{lc}
\toprule
\textbf{Method} & \textbf{Loss} \\
\midrule
LLE  & 0.0484 \\
Laplacian EigenMaps & 0.0652\\
ISOMAP  & 0.0459 \\
t-SNE   & 0.1272 \\
Autoencoder Baseline & 0.0173 \\
\bottomrule
\end{tabular}
\end{table}

\subsection{Locally Linear Embedding Reconstruction }

The LLE implementation operates on flattened 12,288-dimensional image vectors, configured with k=50 nearest neighbors to ensure sufficient local neighborhood coverage for robust weight computation. The manifold embedding projects data to a 256-dimensional latent space, preserving local linear relationships while achieving substantial dimensionality reduction. Our decoder architecture successfully reconstructs images from these LLE coordinates, demonstrating the feasibility of inverse mapping from locally-linear manifold representations to pixel space. 

\begin{figure*}[h!]
    \centering
    \includegraphics[width=\textwidth]{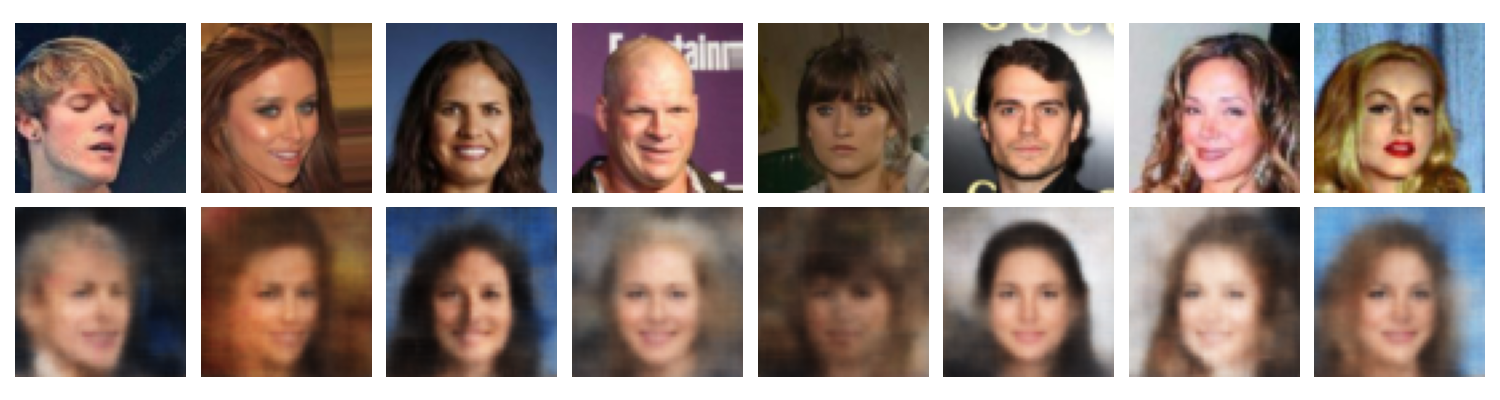}
    \caption{Decoder reconstruction results on the CelebA dataset using LLE-based nonlinear embeddings.}
    
    \label{fig:placeholder}
\end{figure*}

\subsection{Laplacian Eigenmaps Decoder Performance}
The spectral clustering approach employs Laplacian Eigenmaps with a critical architectural modification necessitated by downstream diffusion requirements. To maintain compatibility with our diffusion pipeline, input images are resized to 28×28 grayscale format, yielding 784-dimensional feature vectors. The spectral decomposition preserves the top 784 eigenvectors, with decoder hidden dimensions configured to match this eigenspace dimensionality. This configuration ensures proper alignment between manifold embedding characteristics and decoder input requirements, though the reduced spatial resolution constrains fine-grained reconstruction detail. 
\begin{figure*}[h!]
    \centering
    \includegraphics[width=\textwidth]{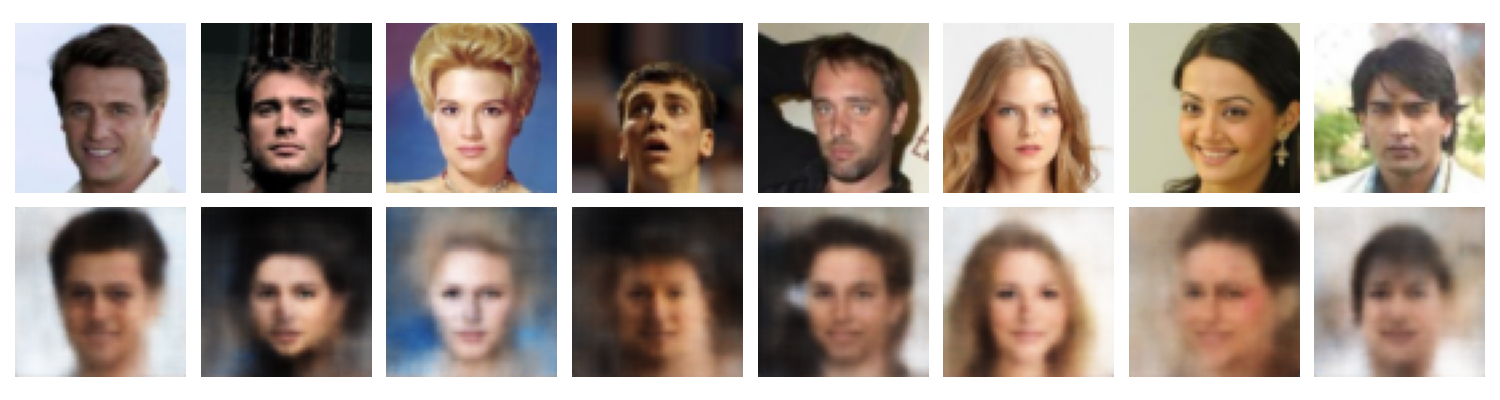}
    \caption{Decoder reconstruction results on the CelebA dataset using Laplacian EigenMaps-based nonlinear embeddings.}
    \label{fig:placeholder}
\end{figure*}

\subsection{ISOMAP Geodesic Reconstruction}

The ISOMAP decoder maintains architectural consistency with the LLE implementation, employing identical network configurations and training protocols. This design choice enables direct comparison between local linear preservation (LLE) and geodesic distance preservation (ISOMAP) while controlling for decoder-specific effects. Both methods operate on the same 25,000-image subset with 64×64 RGB resolution, 256-dimensional latent space, and k=50 neighborhood parameters, isolating the impact of different manifold learning objectives on reconstruction quality. 

\begin{figure*}[h!]
    \centering
    \includegraphics[width=\textwidth]{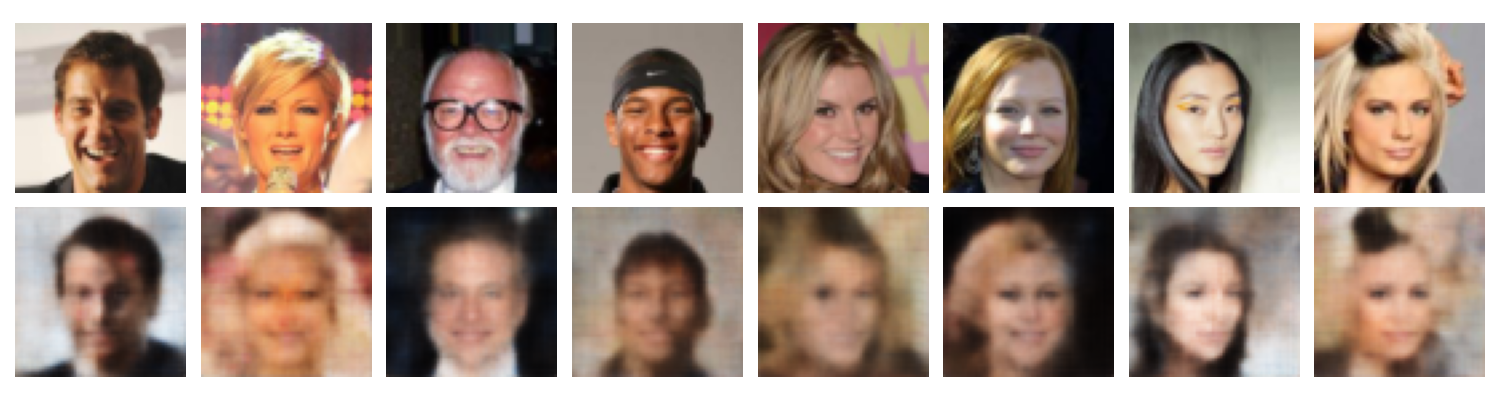}
    \caption{Decoder reconstruction results on the CelebA dataset using ISOMAP-based nonlinear embeddings.}
    \label{fig:placeholder}
\end{figure*}

\subsection{t-SNE Probabilistic Embedding Analysis }

The t-SNE implementation optimizes probabilistic neighborhood relationships using perplexity=30, balancing local structure preservation with global organization over 1000 optimization iterations. Unlike other methods, t-SNE employs a 3-dimensional embedding space, reflecting its primary application in visualization tasks. The Barnes-Hut approximation accelerates computation while maintaining embedding quality. Despite the extremely low-dimensional representation, our decoder architecture adapts to reconstruct full-resolution images from these 3D probabilistic coordinates. 

\begin{figure*}[h!]
    \centering
    \includegraphics[width=\textwidth]{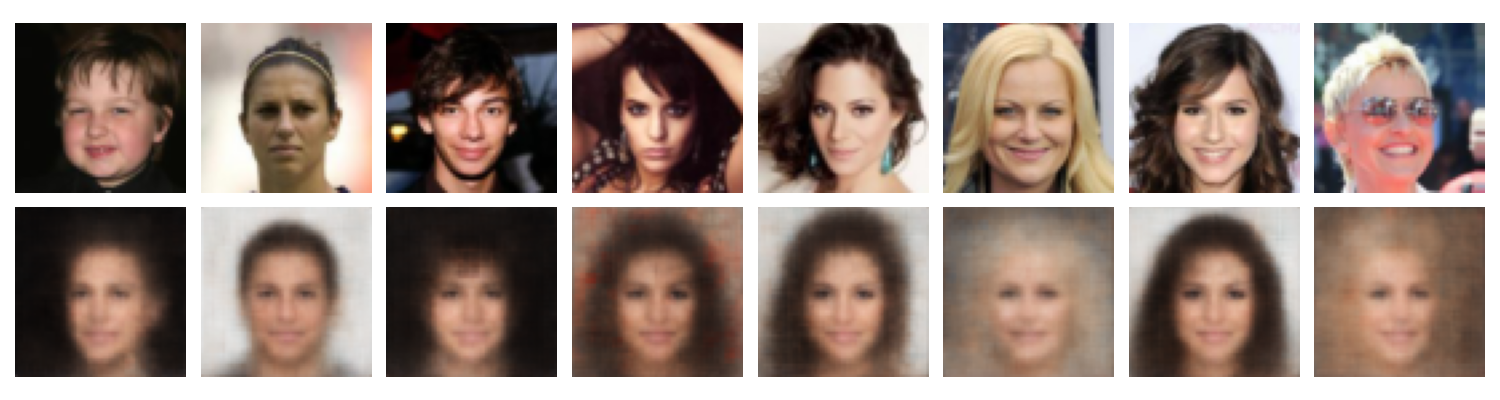}
    \caption{Decoder reconstruction results on the CelebA dataset using TSNE-based nonlinear embeddings.}
    \label{fig:placeholder}
\end{figure*}

\subsection{Autoencoder Baseline Comparison }
The autoencoder baseline establishes performance bounds for joint encoder-decoder optimization, contrasting with the sequential approach necessitated by fixed NLDR embeddings. The encoder employs four convolutional layers with progressive channel expansion (3→32→64→128→256) and spatial compression through 2×2 max pooling, utilizing LeakyReLU activation and batch normalization for stable training dynamics. The decoder mirrors this architecture through transposed convolutions, culminating in tanh activation for normalized output generation. This joint optimization paradigm enables superior reconstruction performance compared to manifold-based approaches, highlighting the fundamental challenge of retrofitting decoders to pre-computed embeddings. 

\begin{figure*}[h!]
    \centering
    \includegraphics[width=\textwidth]{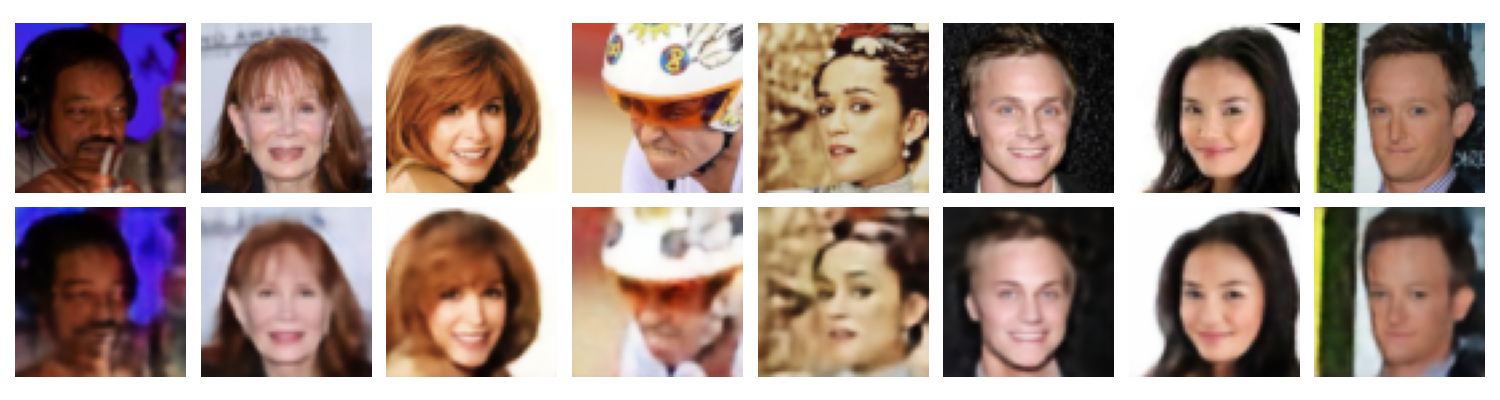}
    \caption{Decoder reconstruction results on the CelebA dataset using an Autoencoder.}
    \label{fig:placeholder}
\end{figure*}

\subsection{Baseline Diffusion Model Performance}

\begin{figure*}[h!]
    \centering
    \includegraphics[width=\textwidth]{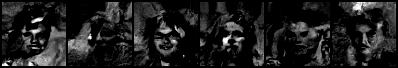}
    \caption{Baseline DDPM results on grayscale CelebA images, illustrating generative quality prior to manifold-based diffusion experiments.}

    \label{fig:placeholder}
\end{figure*}

We establish diffusion model baselines using a standard DDPM implementation on grayscale CelebA images. Initial experiments with the default U-Net architecture yielded suboptimal generation quality, prompting architectural scaling through a five-fold parameter increase. While this modification improved sample fidelity, memory constraints prevented further scaling, limiting achievable quality. The final baseline model trains on 100,000 images at 64×64 grayscale resolution over 1000 diffusion timesteps, providing a reference point for manifold-based generation evaluation.

\subsection{Manifold-Constrained Diffusion Analysis}
Having demonstrated successful reconstruction from individual NLDR embeddings, we proceed to evaluate generative capabilities through manifold-constrained diffusion. This approach performs iterative denoising directly within learned manifold spaces, subsequently projecting generated coordinates through trained decoders to synthesize novel images. The manifold-constrained framework theoretically offers advantages through geometric structure preservation and reduced computational requirements compared to high-dimensional image diffusion. 

However, experimental results reveal significant challenges in this approach. Generated samples exhibit substantial degradation compared to original dataset images, suggesting fundamental limitations in either manifold diffusion implementation or decoder generalization to novel embedding regions. The poor reconstruction quality indicates that current manifold learning techniques may inadequately capture the continuous structures necessary for effective generative modeling, or that decoder training insufficient explores interpolated regions of the embedding space. 
\begin{figure*}[h!]
    \centering
    \includegraphics[width=\textwidth]{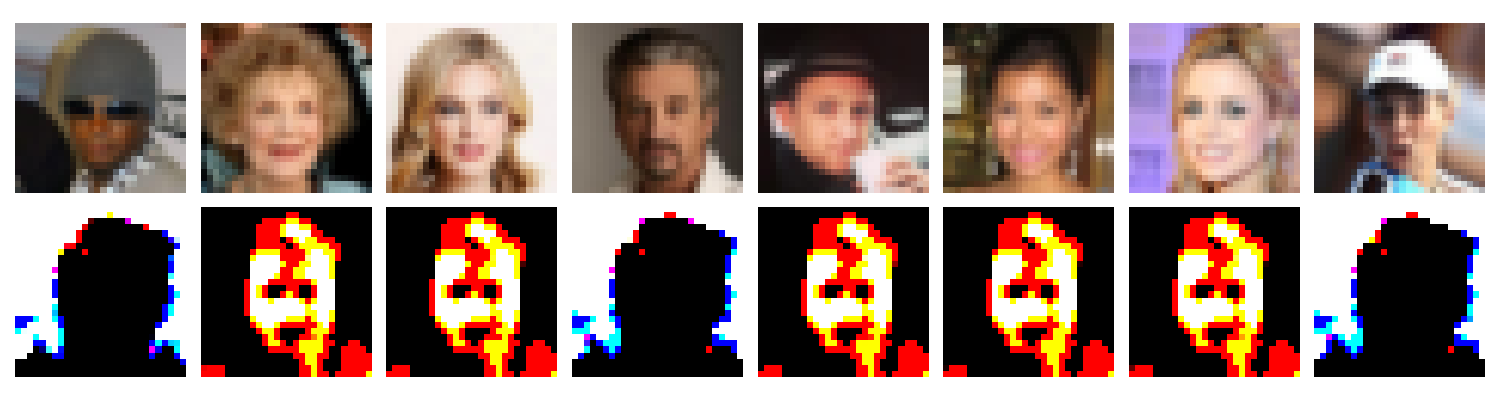}
    \caption{Manifold-constrained diffusion results on NLDR embeddings, illustrating degraded image quality and highlighting the limitations of current manifold and decoder formulations for generative synthesis.}

    \label{fig:placeholder}
\end{figure*}

\section{Discussion and Conclusion}

\subsection{Computational Efficiency and Scalability Considerations}
Contrary to initial expectations that manifold learning would provide computational advantages through dimensionality reduction, our experimental analysis reveals significant computational bottlenecks inherent to traditional NLDR methods. The predominant reliance on eigendecomposition, nearest-neighbor graph construction, and iterative optimization procedures in methods such as LLE, ISOMAP, and Laplacian Eigenmaps necessitates CPU-based computation, creating substantial performance disparities compared to GPU-accelerated neural architectures. Despite extensive efforts to adapt these algorithms for parallel GPU execution, the sequential nature of eigenvalue computations and sparse matrix operations presents fundamental barriers to efficient vectorization. Consequently, while manifold-based diffusion operates in lower-dimensional spaces theoretically reducing computational complexity, the preprocessing overhead for manifold computation substantially outweighs these potential gains. 

\subsection{Reconstruction Quality Assessment and Architectural Limitations }
Our systematic evaluation across four prominent NLDR techniques demonstrates consistent reconstruction quality deficiencies compared to autoencoder baselines when applied to the CelebA facial dataset. This performance gap stems from fundamental architectural differences: autoencoders jointly optimize encoder-decoder mappings through end-to-end gradient descent, enabling learned representations specifically tailored for reconstruction objectives. In contrast, NLDR methods optimize embeddings according to geometric criteria (local linearity, geodesic preservation, spectral properties) without explicit reconstruction considerations, creating a fundamental mismatch between embedding objectives and decoder requirements.

The reconstruction degradation manifests most severely in high-frequency details and texture preservation, suggesting that traditional manifold learning techniques inadequately capture the continuous differentiable structures necessary for pixel-level image synthesis. T-SNE's probabilistic neighborhood optimization, while effective for visualization, proves particularly unsuitable for reconstruction due to its emphasis on cluster separation rather than smooth interpolation properties essential for generative applications. 

\subsection{Manifold-Constrained Diffusion: Challenges and Theoretical Implications }
The poor performance of manifold-constrained diffusion reveals deeper theoretical limitations in applying generative modeling frameworks to non-parametric dimensionality reduction techniques. Our results suggest that traditional NLDR embeddings exhibit insufficient coverage and continuity for effective sample generation, with decoder networks struggling to generalize beyond the discrete manifold coordinates observed during training. The dramatic quality degradation in diffusion-generated samples indicates that learned decoders fail to interpolate meaningfully in regions of the embedding space not explicitly represented in the training data.

This limitation highlights a critical distinction between visualization-oriented manifold learning and generation-oriented representation learning. While NLDR methods excel at preserving local neighborhood structures for analysis tasks, they may not provide the smooth, continuous latent spaces required for high-quality generative modeling. The discrete, often sparse nature of computed embeddings creates substantial challenges for diffusion processes that rely on continuous probability distributions and smooth denoising trajectories. 
\subsection{Broader Implications for Generative Manifold Learning}
These findings illuminate fundamental tensions between geometric structure preservation and generative capability in dimensionality reduction techniques. The superior performance of autoencoders reflects their explicit optimization for bidirectional transformation quality, while traditional NLDR methods prioritize geometric property preservation over reconstruction fidelity. This suggests that effective generative manifold learning may require novel architectures that explicitly balance geometric structure discovery with reconstruction quality during joint optimization. 

\subsection{Future Research Directions}

Several promising avenues emerge from this analysis to address the identified limitations. Hybrid architectures that incorporate geometric regularization terms from manifold learning into autoencoder training objectives could potentially combine the structural insights of NLDR methods with the reconstruction capabilities of neural approaches. Additionally, developing differentiable approximations to traditional manifold learning algorithms would enable end-to-end optimization while preserving geometric properties.

Advanced decoder architectures incorporating attention mechanisms, progressive generation, or adversarial training could improve reconstruction quality from fixed manifold embeddings. Furthermore, investigating manifold-aware diffusion processes that explicitly model the geometric properties of NLDR embeddings, rather than treating them as Euclidean coordinates, may yield improved generation quality.

Finally, addressing computational scalability through specialized hardware implementations, approximation algorithms, or neuromorphic computing approaches could mitigate the CPU bottlenecks that currently limit practical applications of manifold-based generative modeling. These research directions collectively suggest that while current approaches face significant challenges, the fundamental concept of manifold-constrained generation remains promising with appropriate methodological refinements.

\section*{Data and Code Availability}

The code and data used in this study are publicly available on GitHub: \url{https://github.com/kakugri/manifoldDiffusion}.

\bibliography{references}

\begin{thebibliography}{13}
\providecommand{\natexlab}[1]{#1}
\providecommand{\url}[1]{\texttt{#1}}
\expandafter\ifx\csname urlstyle\endcsname\relax
  \providecommand{\doi}[1]{doi: #1}\else
  \providecommand{\doi}{doi: \begingroup \urlstyle{rm}\Url}\fi

\bibitem[Belkin and Niyogi(2003)]{Belkin2003Laplacian}
M.~Belkin and P.~Niyogi.
\newblock Laplacian eigenmaps for dimensionality reduction and data representation.
\newblock \emph{Neural Computation}, 2003.

\bibitem[Bengio et~al.(2003)Bengio, Delalleau, and Le~Roux]{Bengio2003ManifoldParzen}
Y.~Bengio, O.~Delalleau, and N.~Le~Roux.
\newblock Non-local manifold parzen windows.
\newblock In \emph{Advances in Neural Information Processing Systems}, 2003.

\bibitem[De~Bortoli et~al.(2022)De~Bortoli, Mathieu, Hutchinson, Thornton, Teh, and Doucet]{debortoli2022riemannian}
Valentin De~Bortoli, Erwan Mathieu, Mark Hutchinson, James Thornton, Yee~Whye Teh, and Arnaud Doucet.
\newblock Riemannian score-based generative modelling.
\newblock In \emph{Advances in Neural Information Processing Systems}, volume~35, pages 2406--2422, 2022.

\bibitem[Hinton and Salakhutdinov(2006)]{Hinton2006Autoencoder}
G.~E. Hinton and R.~R. Salakhutdinov.
\newblock Reducing the dimensionality of data with neural networks.
\newblock \emph{Science}, 2006.

\bibitem[Ho et~al.(2020)Ho, Jain, and Abbeel]{Ho2020DDPM}
J.~Ho, A.~Jain, and P.~Abbeel.
\newblock Denoising diffusion probabilistic models.
\newblock \emph{arXiv preprint arXiv:2006.11239}, 2020.
\newblock \doi{10.48550/arXiv.2006.11239}.

\bibitem[Ioffe and Szegedy(2015)]{ioffe2015batch}
Sergey Ioffe and Christian Szegedy.
\newblock Batch normalization: Accelerating deep network training by reducing internal covariate shift.
\newblock In \emph{International Conference on Machine Learning}, pages 448--456. PMLR, 2015.

\bibitem[Liu et~al.(2015)Liu, Luo, Wang, and Tang]{liu2015deep}
Ziwei Liu, Ping Luo, Xiaogang Wang, and Xiaoou Tang.
\newblock Deep learning face attributes in the wild.
\newblock In \emph{Proceedings of the IEEE International Conference on Computer Vision}, pages 3730--3738, 2015.

\bibitem[McInnes et~al.(2018)McInnes, Healy, and Melville]{McInnes2018UMAP}
Leland McInnes, John Healy, and James Melville.
\newblock Umap: Uniform manifold approximation and projection for dimension reduction.
\newblock \emph{arXiv preprint arXiv:1802.03426}, 2018.

\bibitem[Pu et~al.(2016)Pu, Gan, Henao, Yuan, Li, Stevens, and Carin]{pu2016variational}
Yelong Pu, Zhe Gan, Ricardo Henao, Xin Yuan, Chunyuan Li, Austin Stevens, and Lawrence Carin.
\newblock Variational autoencoder for deep learning of images, labels and captions.
\newblock In \emph{Advances in Neural Information Processing Systems}, volume~29, 2016.

\bibitem[Roweis and Saul(2000)]{roweis2000lle}
S.~T. Roweis and L.~K. Saul.
\newblock Nonlinear dimensionality reduction by locally linear embedding.
\newblock \emph{Science}, 2000.

\bibitem[Tammina(2019)]{tammina2019transfer}
Sreeram Tammina.
\newblock Transfer learning using vgg-16 with deep convolutional neural network for classifying images.
\newblock \emph{International Journal of Scientific and Research Publications (IJSRP)}, 9\penalty0 (10):\penalty0 143--150, 2019.

\bibitem[Tenenbaum et~al.(2000)Tenenbaum, de~Silva, and Langford]{Tenenbaum2000Isomap}
J.~B. Tenenbaum, V.~de~Silva, and J.~C. Langford.
\newblock A global geometric framework for nonlinear dimensionality reduction.
\newblock \emph{Science}, 290, 2000.

\bibitem[van~der Maaten and Hinton(2008)]{Maaten2008TSNE}
L.~van~der Maaten and G.~Hinton.
\newblock Visualizing data using t-sne.
\newblock \emph{Journal of Machine Learning Research}, 2008.

\end{thebibliography}

\end{document}